# Hierarchical GPT with Congruent Transformers for Multi-Sentence Language Models

**Jihyeon Roh[1]**   **Huiseong Gim[2]**   **Soo-Young Lee[1,2]**

[1] School of Electrical Engineering, Korea Advanced Institute of Science and Technology
[2] KI for Artificial Intelligence, Korea Advanced Institute of Science and Technology
rohleejh@kaist.ac.kr, hsgim@kaist.ac.kr, sy-lee@kaist.ac.kr

## Abstract

We report a GPT-based multi-sentence language model for dialogue generation and document understanding. First, we propose a hierarchical GPT which consists of three blocks, i.e., a sentence encoding block, a sentence generating block, and a sentence decoding block. The sentence encoding and decoding blocks are basically the encoder-decoder blocks of the standard Transformers, which work on each sentence independently. The sentence generating block is inserted between the encoding and decoding blocks, and generates the next sentence embedding vector from the previous sentence embedding vectors. We believe it is the way human make conversation and understand paragraphs and documents. Since each sentence may consist of fewer words, the sentence encoding and decoding Transformers can use much smaller dimensional embedding vectors. Secondly, we note the attention in the Transformers utilizes the inner-product similarity measure. Therefore, to compare the two vectors in the same space, we set the transform matrices for queries and keys to be the same. Otherwise, the similarity concept is incongruent. We report experimental results to show that these two modifications increase the language model performance for tasks with multiple sentences.

## Introduction

Automatic multi-turn dialogue generation and document writing require language models with much longer word sequences. Due to the word-sequence generative nature from the previous dialogues or already-written documents of these tasks, the encoder-decoder seq2seq model (Bahdanau et al., 2015) based on long short-term memory (LSTM; Hochreiter and Schmidhuber 1997) network, a gated recurrent neural network (GRU; Chung et al. 2015), or a Transformer (Vaswani et al. 2017) have been the natural choices. Also, based on the hierarchical nature from characters to multiple sentences via words and sentences (or utterances), hierarchical architectures have potential advantages over the popular one-dimensional architectures. Therefore, hierarchical LSTMs have been widely adopted for natural language processing tasks (Nallapati et al. 2017; Narayan et al. 2018; Zhang et al. 2018). Since recent studies (Vaswani et al. 2017; Devlin et al. 2018; Liu et al. 2019; Radford et al. 2019; Brown et al. 2020) show the Transformer-based models outperform LSTM-based models in many sequence generation tasks, we would like to use a hierarchical Transformer model.

A few hierarchical Transformer models had been reported for document summarization (Zhang, Wei, and Zhou 2019; Liu and Lapata 2019), document classification (Pappagari et al. 2019), and question-answering (Li and Choi 2020) tasks. These models are based on Bidirectional Encoder Representations from Transformers (BERT) which basically utilizes Transformer encoder blocks. On the other hand, the generative pre-training (GPT; Radford et al. 2019; Brown et al. 2020) models utilize Transformer decoder blocks, and have demonstrated the state-of-the-art performance on auto-regressive tasks.

In this paper we report a new autoregressive model (Hierarchical GPT) which takes advantage of the encoder-decode models with a hierarchical architecture. The developed Hierarchical GPT model consists of three blocks, i.e., a sentence (or utterance) encoder block, a document (or dialogue) decoder block, and a sentence (or utterance) decoder lock. While the sentence encoding block incorporates the context of all words in a sentence for better results, the two decoder blocks are auto-regressive suitable for sequence generation. Although XLNet (Yang 2019) is auto-regressive in nature but brings backward regression for better performance, it does not have the hierarchical advantage. Also, due to the hierarchical architecture, both the sentence encoder/decoder and document decoder require much small number of tokens, which results in great computational efficiency. Transformer-XL (Dai et al. 2019) obtained the computational efficiency by regularly segmenting the token sequences and incorporating a method to utilize information in the previous segment. However, the regular segmentation does not match with the hierarchical structure in natural languages.

We also add a constraint to all Transformer blocks. The scaled dot-product attention of Transformers (Vaswani et al. 2017) basically estimates similarities between query and key embedding vectors. At each head this similarity is calculated in a transformed space, i.e., between $\mathbf{W}^q\mathbf{q}$ and $\mathbf{W}^k\mathbf{k}$, with transformation matrices $\mathbf{W}^q$ for query $\mathbf{q}$ and $\mathbf{W}^k$ for key $\mathbf{k}$. We believe the dot-product similarity for both query and key need be evaluated in a same transformed space and the transformation matrices $\mathbf{W}^q$ and $\mathbf{W}^k$ need be the same. This model is called as congruent Transformer.

We report results of language model performance for Pen Tree Bank (PTB) and Text8 datasets. The proposed hierarchical GPT shows better performance than the standard Transformer-based GPT, and the congruent Transformers result in better Perplexity and robustness in network hyper parameters. The best results are obtained by congruent hierarchical GPTs.

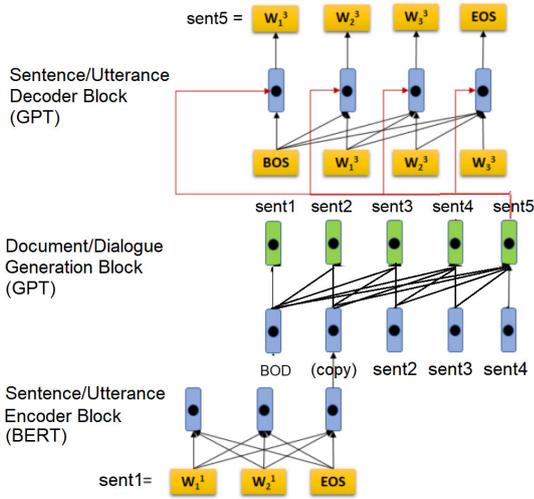

Figure 1. The architecture of the proposed Hierarchical GPT. One Sentence/Utterance Encoder block (BERT) and one Decoder block (GPT) are used for all sentences. At the Sentence/Utterance Encoder block each sentence becomes one training example, and the last output becomes the sentence (or utterance) embedding vector. All these sentence embeddings are fed to the Document/Dialogue Generation block (GPT) with a Begin-Of-Document/Dialogue (BOD) token, and autoregressively learn to generate the next sentence embeddings. At the Sentence/Utterance Decoder block a GPT autoregressively generates word (or token) sequences for each sentence.

## Model Architecture and Training Strategies

### Hierarchical GPT Model

Figure 1 shows network architecture of the proposed Hierarchical GPT, which consists of three Transformer-based blocks. For simplicity, we only show the Sentence/Utterance Encoder block for the first sentence and the Sentence/Utterance Decoder block for the fifth sentence, respectively.

At the Sentence/Utterance Encoder block a BERT learns to encode a sentence (or utterance) into an embedding vector from a word (or token) sequence. Each sentence becomes an independent training sample, and the required number of words (or tokens) becomes much smaller. For standard one-dimensional BERT a big word sequence from several sentences need be inputted to the Transformer.

All these sentence embedding vectors become the inputs to the Document/Dialogue Generation block with a learnable Begin-Of-Document/Dialogue (BOD) token embedding vector. Here a GPT learns to generate the next sentence (or utterance) embedding vectors from previous sentence embedding vectors. The number of sentences for each document (or dialogue) also becomes much smaller than that of words (or tokens) in the document.

At the Sentence/Utterance Decoder block, a GPT learns to generate a word or token sequence for each generated sentence embedding vector.

Each block may consist of several Transformers. Also, the hierarchical architecture may extend into more levels. For example, one may adopt character level tokens at the lowest inputs, and build up progressively more complex embeddings through subwords, words, phrases, sentences, paragraphs, sections, chapter, and books.

For the training one may select one of two different strategies. At the first strategy the Sentence/Utterance Encoder and Decoder blocks are directly connected without the Document/Dialogue Generation block, and trained first as a simple seq2seq encoder-decoder network. Then, using these learned sentence embedding vectors, the Document/Dialogue Generation block is trained. At the second strategy all the three blocks in Figure 1 may be trained simultaneously with random initialization or the results from the first strategy.

### Congruent Transformer Model

Although the standard Transformers utilize separate transformation matrices $\mathbf{W}^q$ and $\mathbf{W}^k$ for query $\mathbf{q}$ and key $\mathbf{k}$, respectively, we believe the dot-product similarity measure is good only between two vectors in a same space.

As shown in Figure 2, a vector-matrix multiplication $\mathbf{Wx}$ linearly transforms a vector $\mathbf{x}$ into another space $\mathbf{x}'$. In general, the linear transformation consists of three transforms, i.e., elongation, tilting, and rotation, and an arbitrary transform matrix can be factorized into 3 matrices, i.e., an elongation matrix $\mathbf{E}$, a tilting matrix $\mathbf{T}$, and a rotation matrix $\mathbf{R}$.

For an example of two-dimensional space, any matrix **W** can be factorized as

**W=RTE**,

$$\mathbf{E}=\begin{bmatrix} e1 & 0 \\ 0 & e2 \end{bmatrix}, \quad \mathbf{T} = \begin{bmatrix} 1-c & 0 \\ 0 & 1 \end{bmatrix}, \quad \mathbf{R} = \begin{bmatrix} \cos\theta & -\sin\theta \\ \sin\theta & \cos\theta \end{bmatrix}.$$

Here, $e1$ and $e2$ are elongation factors of x1 and x2 axes, respectively, c is a tilting coefficient, and $\theta$ is a rotation angle. Therefore, different $\mathbf{W^q}$ and $\mathbf{W^k}$ matrices result in different transformation, i.e., different elongation factor, different tilting scale, and/or different rotation angle. Calculating similarity of two vectors in two different transformed space become incongruent and has no physical meaning.

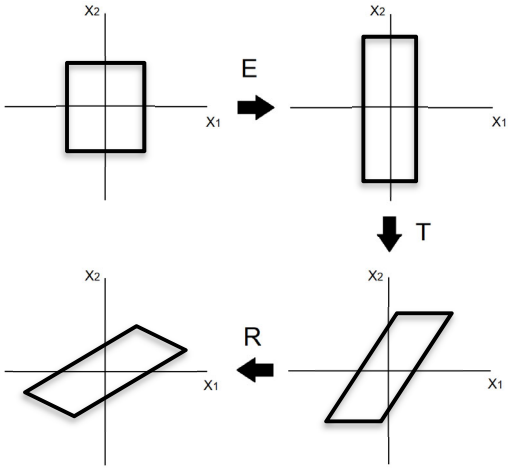

Figure 2. A linear transform matrix **W** maps a vector **x** into another space by **x'=Wx**. The transform matrix **W** may be decomposed into an elongation matrix **E**, a tilting matrix **T**, and a rotation matrix **R**.

Therefore, for each head of the Transformers we enforce a same transformation matrix **W** for both query **q** and key **k**, i.e., $\mathbf{W^q} = \mathbf{W^k}$. With the smaller number of learnable parameters, it will help to reduce overfitting for better generalization. If the word embedding vectors of big vocabulary corpus dominate the number of learnable parameters, this effect may not be important. However, we may reduce the number of learnable parameters with character-level tokens. Also, it may be possible to utilize pre-trained word embedding vectors and adapt the others only.

## Datasets and Experimental Settings

To validate our method, we used a small but popular language modeling benchmark dataset, i.e., the Penn Treebank (PTB) dataset (Marcus et al. 1993). The Penn Treebank is a small dataset with about 10K vocabulary and 1M words, and is a subset of the Wall Street Journal (WSJ) dataset that contains news articles with business-oriented topics. Since we are trying to figure out the effects of the hierarchical architecture and congruent Transformers and also due to the limitation of our computing facilities, we use a small dataset with several different numbers of learnable parameters.

In this paper we report results from the second strategy. All the three blocks in Figure 1 were trained simultaneously with random initialization. For each network several experiments were conducted and the learning convergence was confirmed.

## Experimental Results

First, we show results of GPTs based on the Transformers without and with the congruent condition, i.e., $\mathbf{W^q} = \mathbf{W^k}$ for each head for the small PTB dataset. The embedding dimension was set to 512 and the number of heads is 8 for all these experiments. However, the number of Transformer layers varies from 4 to 15.

Table 1. Perplexity (PPL) values of the standard GPTs for several different numbers of layers. Here, each Transformer has 8 heads with 512 dimensional embedding vectors.

| Number of Layers | 4 | 6 | 8 | 10 | 12 | 15 |
|---|---|---|---|---|---|---|
| Number of Parameters | 20.3M | 26.6M | 32.9M | 39.2M | 45.5M | 55.0M |
| Train PPL | 33.6 | 31.2 | 29.1 | 33.5 | 31.6 | 31.4 |
| Valid PPL | 64.6 | 62.1 | 61.7 | 61.8 | 61.3 | 63.3 |
| Test PPL | 54.9 | 52.9 | 52.4 | 52.8 | 52.4 | 54.0 |

Table 2. Perplexity (PPL) values of GPTs with congruent Transformers for several different numbers of layers. Here, each Transformer has 8 heads with 512 dimensional embedding vectors.

| Number of Layers | 4 | 6 | 8 | 10 | 12 | 15 |
|---|---|---|---|---|---|---|
| Number of Parameters | 19.3M | 25.0M | 30.8M | 36.6M | 42.4M | 51.0M |
| Train PPL | 29.5 | 27.8 | 29.0 | 31.9 | 30.7 | 30.7 |
| Valid PPL | 58.2 | 57.8 | 58.7 | 58.9 | 59.4 | 61.7 |
| Test PPL | 50.5 | 50.3 | 50.7 | 51.1 | 51.5 | 53.2 |

As shown in Tables 1 and 2, GPTs with the congruent Transformers consistently outperform GPTs with the standard Transformers. More importantly, the congruent Transformers greatly reduce the performance variations from different numbers of layers. For example, all the PPL values for the training, validation, and test datasets do not change much for the GPTs with 6 to 12 congruent Transformer layers. It demonstrates that the proposed congruent constraint on the transformation matrices works well to improve robustness.

Secondly, in Table 3 we show results from the Hierarchical GPTs with several numbers of Transformer layers. In these experiments we wanted to find the blocks requiring more layers at a given number of total layers. Although only one experiment was performed for each number combination of Transformer layers in the row, it consistently shows that the cases with more layers at the Sentence Decoder block (bold fonts) outperform the others. It may come from the fact that the decoders conduct much complex functions than the encoders. Therefore, for the latter experiments, we only use these cases to integrate congruent Transformers into Hierarchical GPTs.

Table 3. Perplexity (PPL) values of Hierarchical GPTs with several numbers of layers. Here, each Transformer has 8 heads with 512 embedding dimension.

| Number of Layers | | | | PPL | | |
|---|---|---|---|---|---|---|
| Total | Sentence Encoder | Doc Encoder | Sent Decoder | Train Data | Valid Data | Test Data |
| 3 | 1 | 1 | 1 | 43.3 | 68.8 | 58.4 |
| 4 | 1 | 1 | 2 | 33.7 | 62.1 | 53.0 |
| 5 | **1** | **1** | **3** | 29.9 | 60.1 | **51.5** |
|   | 2 | 2 | 1 | 37.9 | 67.3 | 56.9 |
| 6 | **1** | **1** | **4** | 31.4 | 58.8 | **50.5** |
|   | 2 | 2 | 2 | 40.6 | 61.8 | 52.6 |
| 7 | **2** | **2** | **3** | 29.2 | 60.0 | **50.8** |
|   | 3 | 3 | 1 | 38.6 | 66.6 | 56.1 |
| 8 | **2** | **2** | **4** | 30.5 | 59.0 | **50.5** |
|   | 3 | 3 | 2 | 32.8 | 62.1 | 52.7 |
| 9 | **3** | **3** | **3** | 34.7 | 60.7 | **51.6** |
|   | 4 | 4 | 2 | 41.5 | 67.5 | 57.1 |
| 10 | **3** | **3** | **4** | 31.0 | 59.5 | **51.1** |
|    | 4 | 4 | 2 | 34.3 | 62.1 | 53.0 |
| 11 | 4 | 4 | 3 | 32.5 | 62.6 | 53.3 |
| 12 | 4 | 4 | 4 | 32.1 | 60.1 | 51.6 |
| 13 | 5 | 5 | 3 | 34.5 | 62.0 | 53.4 |
| 14 | 5 | 5 | 4 | 32.0 | 62.5 | 53.2 |
| 15 | **5** | **5** | **5** | 32.1 | 62.2 | **53.4** |
|    | 6 | 6 | 3 | 33.2 | 62.1 | 53.5 |

As shown in Tables 4 and 5, Hierarchical GPTs with the congruent Transformers consistently outperform Hierarchical GPTs with the standard Transformers. Again, the Hierarchical GPTs with congruent Transformers greatly reduce the performance variations from different numbers of layers. For example, the test data PPL values do not change much with 6 to 12 congruent Transformer layers. It demonstrates that the proposed congruent constraint on the transformation matrices works well with the Hierarchical GPTs.

Table 4. Perplexity (PPL) values of Hierarchical GPTs with the standard Transformers for several different numbers of layers. Here, each Transformer has 8 heads with 512 dimensional embedding vectors.

| Number of Layers | 4 | 6 | 8 | 10 | 12 | 15 |
|---|---|---|---|---|---|---|
| Number of Parameters | 20.3M | 26.6M | 32.9M | 39.2M | 45.5M | 55.0M |
| Train PPL | 33.7 | 31.4 | 30.5 | 31.0 | 32.1 | 32.1 |
| Valid PPL | 62.1 | 58.8 | 59.0 | 59.5 | 60.1 | 62.2 |
| Test PPL | 53.0 | **50.5** | **50.5** | 51.1 | 51.6 | 53.4 |

Table 5. Perplexity (PPL) values of Hierarchical GPTs with congruent Transformers for several different numbers of layers. Here, each Transformer has 8 heads with 512 dimensional embedding vectors.

| Number of Layers | 4 | 6 | 8 | 10 | 12 | 15 |
|---|---|---|---|---|---|---|
| Number of Parameters | 19.3M | 25.0M | 30.8M | 36.6M | 42.4M | 51.0M |
| Train PPL | 32.8 | 30.5 | 31.0 | 29.5 | 29.4 | 31.1 |
| Valid PPL | 61.3 | 58.2 | 58.6 | 58.6 | 59.1 | 59.5 |
| Test PPL | 52.8 | **50.4** | **50.4** | **50.5** | 50.9 | 51.2 |

Figure 3 shows the performance improvements from the congruent condition of the Transformers. The PPL differences between non-hierarchical GPTs of the standard and congruent Transformers are bigger than those of Hierarchical GPTs. However, the differences become smaller with Hierarchical GPTs. The performance of Hierarchical GPTs themselves are improved already, and the congruent Transformers may have less room for improvement. However, for Hierarchical GPTs the differences become much more eminent for larger networks with more Transformer layers. Since the larger networks have more likely to have overfitting problems, it also agrees with our claims that the congruent condition improves robustness by reducing overfitting.

Several improvements are already under planned. To take full advantage of the Hierarchical architecture we need much larger datasets with longer sentence sequences or longer-turn dialogues. We are now looking for larger document database such as WikiText-103 dataset. Also, we are collecting multi-turn dialogue dataset by ourselves. Another big improvement may come from higher-level hierarchy. For example, we plan to extend into three or more hierarchical levels from character-to-syllables, syllable-to-words, words-to-phrases, phrases-to-sentences, sentences-to-paragraphs, and paragraphs-to-Documents.

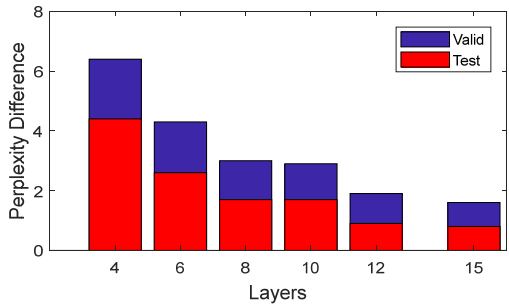

(a)

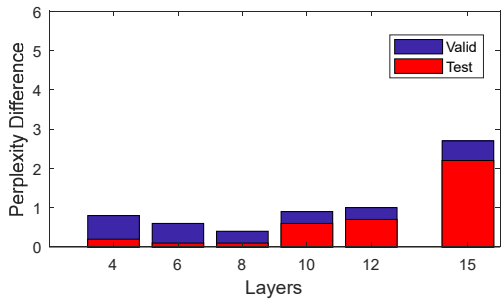

(b)

Figure 3. PPL differences between the standard and congruent Transformers for the validation and test data. (a) the standard GPTs; (b) Hierarchical GPTs.

## Conclusions

In this paper we have proposed a new Hierarchical GPTs and also congruent condition on the transformation matrices $\mathbf{W^q}$ and $\mathbf{W^k}$ for the dot-product similarity estimation. By making experiments with diverse network parameters from 3 Transformer layers to 15 layers for a small Pen Treebank dataset, we demonstrated that the hierarchical GPT architecture and congruent condition result in better performance and robustness over network parameters. Further experiments are in progress for medium and bigger datasets in the future.

# References

Bahdanau, D.; Cho, K.; and Bengio, Y. 2015. Neural machine translation by jointly learning to align and translate. In *Proceedings of the 3rd International Conference on Learning Representations*, San Diego, California.

Brown, T.B.; Mann, B.; Ryder, N.; Subbiah, M.; Kaplan, J.; et al. 2020. Language Models are Few-Shot Learners, *arXiv:2005.14165*.

Chung, J.; Gulcehre, C.; Cho, K.; and Bengio, Y. 2015. Gated feedback recurrent neural networks. In *Proceedings of the 32nd International Conference on Machine Learning*, Lille, France.

Dai Z.; Yang Z.; Yang Y.; Carnonell, J.; Le, Q.V.; and Salakhutdinov, R. Transformer-XL: Attentive language models beyond a fixed-length context. *arXiv: 1901.02860.*

Hochreiter, S, and Schmidhuber, J. 1997. Long short-term memory. *Neural computation* 9(8):1735–1780.

Li, C. and Choi J.D. 2020. Transformers to learn hierarchical contexts in multiparty dialogue for span-based question answering. *arXiv:2004.03561*.

Liu, Y. and Lapata, M. 2019. Hierarchical Transformers for Multi-Document Summarization. *arXiv:1905.13164*.

Liu, Y.; Ott, M.; Goyal, N.; Du, J.; Joshi, M.; Chen, D.; Levy, O.; Lewis, M.; Zettlemoyer, L.; and Stoyanov, V. 2019. RoBERTa: Arobustly optimized BERT pretraining approach. *arXiv:1907.11692*.

Marcus, M.P.; Marcinkiewicz, M.A.; and Santorini, B. 1993. Building a large annotated corpus of English: The Penn Treebank. *Computational linguistics* 19: 313–330.

Nallapati, R.; Zhou, B.; Gulcehre, C.; and Xiang, B. 2016. Abstractive text summarization using sequence-to-sequence rnns and beyond. *arXiv:1602.06023*.

Narayan, S.; Cohen, S.B.; and Lapata, M. 2018. Ranking sentences for extractive summarization with reinforcement learning. In *Proceedings of the 2018 Conference of the North American Chapter of the Association for Computational Linguistics: Human Language Technologies, Volume 1 (Long Papers)*, 1747–1759, New Orleans, Louisiana.

Pappagari, T.; Zelasko, P.; Villala, J.; Carmiel, Y.; and Dehak, N. 2019. Hierarchical Transformers for long document classication. In *Proceedings of IEEE Automatic Speech Recognition and Understanding Workshop*, Sentosa, Singapore.

Radford, A.; Wu, J.; Child, R.; Luan, D.; Amodei, D.; and Sutskever, I. 2019. Language models are unsupervised multitask learners. Technical Report, OpenAI.

Vaswani, A.; Shazeer, N.; Parmar, N.; Uszkoreit, J.; Jones, L.; Gomez, A.N.; Kaiser, L.; and Polosukhin, I. 2017. Attention is all you need. In *Advances in Neural Information Processing Systems*, 5998–6008.

Yang, Z.; Dai, Z.; Yang, Y.; Carbonell, J.; Salakhutdinov, R.; and Le Q.V. 2019, XLNet: Generalized autoregressive pretraining for language understanding. *Neural Information Processing Systems*, Vancouver, Canada.

Zhang, X.; Wei, F.; and Zhou, M. 2019. HIBERT: Document level ore-training of hierarchical bidirectional Transformers for document summarization. *arXiv:1905.06566*.